\documentclass[10pt,twocolumn,letterpaper]{article}

\usepackage{wacv}              

\usepackage{graphicx}
\usepackage{amsmath}
\usepackage{amssymb}
\usepackage{bbm}
\usepackage{booktabs}
\usepackage{multirow,multicol}

\usepackage[pagebackref,breaklinks,colorlinks]{hyperref}

\usepackage[capitalize]{cleveref}
\crefname{section}{Sec.}{Secs.}
\Crefname{section}{Section}{Sections}
\Crefname{table}{Table}{Tables}
\crefname{table}{Tab.}{Tabs.}

\def\wacvPaperID{****} 
\def\confName{WACV}
\def\confYear{2024}

\begin{document}

\title{HIDRO-VQA: High Dynamic Range Oracle for Video Quality Assessment
}

\author{
    Shreshth Saini*, Avinab Saha*, Alan C. Bovik\\
    Laboratory of Image and Video Engineering \\
    The University of Texas at Austin\\
    {\tt\small saini.2@utexas.edu, avinab.saha@utexas.edu, bovik@ece.utexas.edu}
    }

\maketitle

\begin{abstract}
   We introduce HIDRO-VQA, a no-reference (NR) video quality assessment model designed to provide precise quality evaluations of High Dynamic Range (HDR) videos. HDR videos exhibit a broader spectrum of luminance, detail, and color than Standard Dynamic Range (SDR) videos. As HDR content becomes increasingly popular, there is a growing demand for video quality assessment (VQA) algorithms that effectively address distortions unique to HDR content. To address this challenge, we propose a self-supervised contrastive fine-tuning approach to transfer quality-aware features from the SDR to the HDR domain, utilizing unlabeled HDR videos. Our findings demonstrate that self-supervised pre-trained neural networks on SDR content can be further fine-tuned in a self-supervised setting using limited unlabeled HDR videos to achieve state-of-the-art performance on the only publicly available VQA database for HDR content, the LIVE-HDR VQA database. Moreover, our algorithm can be extended to the Full Reference VQA setting, also achieving state-of-the-art performance. Our code is available publicly at \href{https://github.com/avinabsaha/HIDRO-VQA}{https://github.com/avinabsaha/HIDRO-VQA}.

\end{abstract}

{\let\thefootnote\relax\footnote{{This work was supported by the National Science Foundation AI Institute for Foundations of Machine Learning (IFML) under Grant 2019844. *Shreshth Saini and Avinab Saha contributed equally to the work.}}}

\section{Introduction}
\label{sec:intro}
Modern displays are able to present High Dynamic Range (HDR) videos, representing wider ranges of brightness and colors than Standard Dynamic Range (SDR). In this way, HDR videos can deliver more realistic viewing experiences. 

\begin{figure}
  \centering
  \includegraphics[width=\columnwidth]{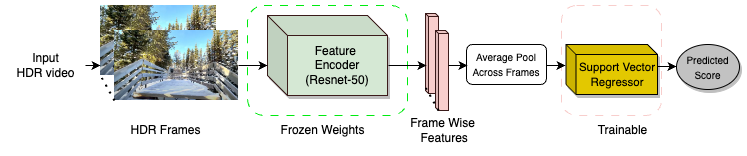}
  \caption{ HIDRO-VQA score prediction using the contrastive Fine Tuned encoder. The encoder is frozen while the regressor learns to map video representations to quality predictions.}
  \label{fig:inference}
\end{figure}  

Display luminance measures the amount of light passing through a specific area, also known as candela per unit area or nits \cite{nits}. The ITU BT.709 \cite{bt.709} video standards were designed for Cathode Ray Tube (CRT) displays, suggesting the use of BT.709 transfer function (commonly known as gamma curve), Rec. 709 color space, and a luminance range of 0.1 $cd/m^2$ to 100 $cd/m^2$. It is worth noting that the human eye can perceive a much wider range of luminances, from $10^{-6}$ $cd/m^2$ to $10^8$ $cd/m^2$ \cite{eye_luma}. 
ITU BT.2100 \cite{bt.2100} introduced HDR-TV by recommending the use of the perceptual quantizer (PQ) or hybrid log–gamma (HLG) transfer functions \cite{hlg,pq}, along with Rec. 2020 color space (wide color gamut), and a nominal peak luminance of 1,000 $cd/m^2$ or more and the black level of 0.005 $cd/m^2$ or less. PQ and HLG systems aim to transform the original scene light into a display-ready representation, better preserving the vision of the artist creators. 

The surge in HDR's adoption and the influx of high-caliber HDR content have significantly elevated viewers' satisfaction ratio \cite{itu-hdrstudy}. The demand for HDR content 
brings a unique challenge, necessitating amplified compression and innovative processing techniques for HDR. The pivotal role of Video Quality Assessment (VQA) models in ensuring optimal HDR video quality during transmission cannot be overstated. Contemporary VQA models play a decisive role in automatically enhancing bitrate determinations on diverse scales in commercial settings. However, existing algorithms work best for SDR content and often fail while delivering HDR content. 
NR-VQA has become a core part of the video infrastructure for streaming and media platforms such as Netflix, YouTube, Instagram, TikTok, X-platform, etc., which enables the streaming of both professionally created content and User Generated Content (UGC). VQA models are essential to objectively predict and control the quality of video content and to help in control the users' Quality of Experience (QoE) \cite{QoE}. Though effective NR-VQA models for SDR content are available \cite{vqa-journey}, their direct application to HDR content consistently fails, as shown in \cite{hdrchipqa}. 

Very little work has been done in the field of HDR-VQA as compared to SDR-VQA, and one major factor contributing to this is the lack of a large-scale dataset. Over the years, a lot of effort has been put into gathering generic SDR-VQA databases, such as  CVD2014 \cite{CVD2014}, LIVE-Qualcomm \cite{live-qualcomm}, LIVQ-VQC \cite{live-vqc}, KoNViD-1k \cite{knovid-1k}, LSVQ \cite{lsvq}, YouTube-UGC \cite{yt-ugc}, and DVL2021 \cite{dvl2021} as well as domain-specific SDR VQA databases focusing on High Frame Rate videos \cite{ythfr}, Live Streaming \cite{livestream}, Cloud Gaming \cite{livemcg} etc. All these existing VQA datasets are small in scale; thus, the field of VQA generally suffers due to limited labeled datasets. At the same time, the availability of large-scale unlabelled videos enables unsupervised or self-supervised approaches. Whereas, in the case of HDR-VQA, there is a very limited availability of HDR videos, and until recently, there has been little work done to create a large-scale labeled HDR dataset.

\subsection{Relevance and Contribution}

With the recent advancement in Deep Learning (DL) methods, many data-driven quality assessment models have been developed \cite{rapique,gamival,musiq}. With the lack of large-scale labeled quality assessment datasets, it is still extremely challenging to train DL-based methods in a supervised manner. Methods like CONTRIQUE \cite{contrique} and Re-IQA \cite{reiqa} proposed exploiting the abundant unlabeled images to learn the quality-aware features without any quality score using contrastive learning. 
No such work has been done in the domain of HDR NR-VQA, largely because of the lack of publicly available HDR videos. 

The market penetration of High Dynamic Range (HDR) capable devices, such as the latest iterations of Apple's iPhones \cite{iphone}, has seen a significant uptick in the past few years. The creator economy, thriving more than ever, has ushered in a wave of high-quality video content, including HDR videos, enriching the digital media ecosystem. This paradigm shift has motivated prominent online platforms like YouTube, TikTok, Meta, Instagram, etc., to improve their infrastructure to host HDR content. Despite this, obtaining quality scores for large-scale HDR content poses a formidable challenge, entailing substantial financial and labor investments. 
To this end, following the philosophy of contrastive learning, 
we propose a self-supervised learning approach for the HDR NR-VQA task. To the best of our knowledge, this work is the first attempt to use self-supervised learning-based approaches for the  HDR NR-VQA tasks on unlabeled HDR videos. We refer to the new model as \textbf{HI}gh \textbf{D}ynamic \textbf{R}ange \textbf{O}racle for \textbf{V}ideo \textbf{Q}uality \textbf{A}ssessment (HIDRO-VQA).
Our contributions are as follows: 

\begin{itemize}
\itemsep0em 
    \item The first contrastive learning-based approach for HDR NR-VQA task leveraging unlabeled HDR videos. 
    \item Our proposed model achieves state-of-the-art performance on the LIVE-HDR \cite{hdr-live} benchmark and outperforms previous HDR VQA algorithms by a large margin. Figure \ref{fig:inference} shows the overview of our proposed model.
\end{itemize}

The remainder of the paper is organized as follows. Section \ref{related} discusses a brief overview of the literature relating to VQA models and self-supervised learning. Section \ref{method} provides details of the proposed model and data pre-processing strategies. Section \ref{experiment} and \ref{ablation} provide experimental results and ablation study, respectively. We also discuss extending our HIDRO-VQA model to FR-VQA in Section \ref{FR-VQA}. Finally, Section \ref{future} concludes the paper by summarizing the contributions of the paper and discussing avenues for future work.

\section{Related Work}\label{related}

VQA models can be broadly divided into hand-crafted feature extraction models and DL-based feature extraction models. Hand-crafted feature extraction model are often training-free and are usually limited in their generalization ability. Learning-based approaches require large-scale labeled datasets but often generalize well on diverse content sets. The NR-VQA task is more challenging, given the enormous range of time-varying distortion combinations that occur on videos. NR-IQA methods can be applied to videos frame-by-frame to estimate the quality scores and then pooled across the temporal dimension, but this often excludes modeling of temporal distortions. To our knowledge, no DL-based HDR NR-VQA method exists that achieves high correlations against human subjective quality scores.


The prevailing design approach behind numerous models involves a dedicated feature extraction framework followed by a regressor to map features to quality scores. In classical models, artifact modeling facilitates feature extraction. A popular example of this approach are models that use Natural Scene Statistics (NSS) or Nature Video Statistics (NVS). NSS models extract features from a transform domain, wherein deviation from expected statistical regularities due to distortions lead to quality estimators. Examples of NSS models include DIIVINE \cite{DIIVINE} with its use of steerable pyramids, V-BLIINDS \cite{vbliinds}, which uses discrete cosine transform coefficients of frame differences, BRISQUE \cite{brisque}, and the unsupervised model NIQE \cite{NIQE} which leverages mean subtracted contrast normalized (MSCN) coefficients of luma to acquire quality-aware features. In CORNIA \cite{CORNIA} and HOSA \cite{hosa}, visual codebooks crafted from local patches are used to obtain quality-aware features. HIGRADE \cite{higrade} an IQA exploits the statistical patterns of the gradient and log-derivative of each channel in CIELAB \cite{CIE} color space. ChipQA \cite{chipqa} models the statistics of space-time chips, which are highly localized space-time slices of MSCN frames. TLVQM \cite{tlvqm} uses several hand-designed, low-complexity features for most recurring distortions like blur, blockiness, motion artifacts, jerkiness, interlacing, etc. High-complexity features (HCF) were sub-sampled at 1 Hz to capture sharpness, blockiness, noise, color, and contrast in CIELAB space. All these hand-crafted feature extraction methods perform well on a limited number of synthetic distortions, but they often fail on real-world distortions. In the experimental section, following HDR-ChipQA \cite{hdrchipqa}, we re-implemented methods that use CIELAB color space to use HDR CIELAB \cite{hdrcie} designed for HDR content.

DL-based models aim to extract semantic and quality-aware features using specialized model architecture, loss functions, or training strategies. Most DL-based IQA approaches use ImageNet \cite{imagenet} pre-trained models and then fine-tune them for the quality assessment task. RAPIQUE \cite{rapique} was designed for the SDR User Generated Content (UGC) VQA task; it combines NVS features and deep CNN features pooled over time. Similar to RAPIQUE, the authors of \cite{gamival} proposed GAMIVAL, which uses a combination of a CNN and neurostatistical NVS features for VQA of SDR gaming videos. 

Minimal exploration has been done on the HDR NR-VQA problem. HDR-BVQM \cite{hdr-bvqm} uses BRISQUE \cite{brisque} features, the log-derivative features defined from HIGRADE \cite{higrade}, and the temporal features from V-BLIINDS \cite{vbliinds}. The HDR-BVQM \cite{hdr-bvqm} design measures statistical consistency only on SDR videos, and thus, we do not treat it as an HDR NR-VQA algorithm. NorVDPNet \cite{norvdpnet}, an HDR-specific method, uses a CNN network trained on proxy quality scores from HDR-VDP \cite{hdrvdp} between reference and distorted image pairs. In HDR-ChipQA \cite{hdrchipqa}, the authors extend ChipQA \cite{chipqa} by adding HDR-specific features. HDR-ChipQA \cite{hdrchipqa} applies non-linearity on the luma values of each HDR frame of a video and extracts the same NVS features as in ChipQA.

Self-supervised learning is directed toward obtaining representations from unlabeled data. This is achieved by tapping into the existing structural information in the image data. Recent SOTA methods use auxiliary tasks that do not require labeled datasets. This includes tasks like rotation prediction \cite{rotation_contrastive}, transitioning between grayscale and color images \cite{zhang2016colorful, larsson2017colorization}, and inpainting \cite{pathak2016context}. In the context of quality assessment, discrimination of distortion types and levels can be utilized as a self-supervision task. CONTRIQUE proposed the use of contrastive learning to exploit large-scale unlabeled synthetic and authentically distorted image databases. Following the same path, Re-IQA also uses contrastive learning but for both quality and content-related features. CONVIQT is another well-recognized VQA model that leverages the pretrained CONTRIQUE's model for spatial feature extraction and further trains a GRU \cite{gru} model to extract temporal quality-aware features in a self-supervised setting.

\begin{figure*}
  \centering
  \includegraphics[width=0.9\textwidth]{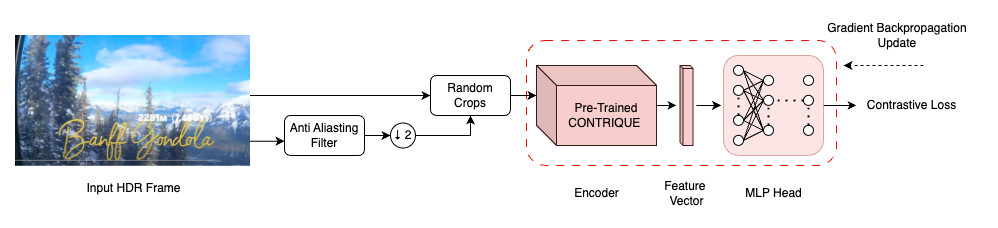}
  \caption{Illustration of fine-tuning pipeline of HIDRO-VQA.}
  \label{fig:finetune_HDR.png}
\end{figure*}

\section{Method : HIDRO-VQA}\label{method}

Our objective is to acquire low-level quality-aware features that accurately characterize HDR videos. Utilizing self-supervised learning, we can achieve this without relying on specific quality scores for HDR videos. We aim to leverage a pre-trained SDR quality-aware model, which has been trained on extensive and diverse datasets, then fine-tune it using a limited collection of HDR videos in a self-supervised setting to obtain final quality-aware HDR representations. Our initial investigation, as demonstrated in Table \ref{tab:nrvqatable}, reveals that self-supervised pre-trained models like Re-IQA, CONTRIQUE, and CONVIQT exhibit strong generalization capabilities on the LIVE-HDR database, even though they were never trained on HDR database. Among these, we deployed CONTRIQUE as our SDR pre-trained model due to its use of a straightforward single-backbone, which contrasts with Re-IQA and CONVIQT, which use multiple sub-models. \\
\indent Much like CONTRIQUE, our approach involves the acquisition of 4096-dimensional feature vectors for any given input video. This vector is derived by extracting frame-level features and averaging them across all frames. Our method diverges from HDR-ChipQA by eliminating the need for NSS-inspired feature extraction. In contrast to CONVIQT, we keep the process simple by abstaining from temporal transformations.  We ensure our model attains a grasp of perceptual distortion features commonly associated with HDR content through our data preparation and fine-tuning strategies. Our HDR quality-aware fine-tuning procedure is illustrated in Figure \ref{fig:finetune_HDR.png} and will be discussed in detail in Section \ref{ssec:finetune}.

\begin{figure*}
  \centering
  \includegraphics[width=0.85\textwidth]{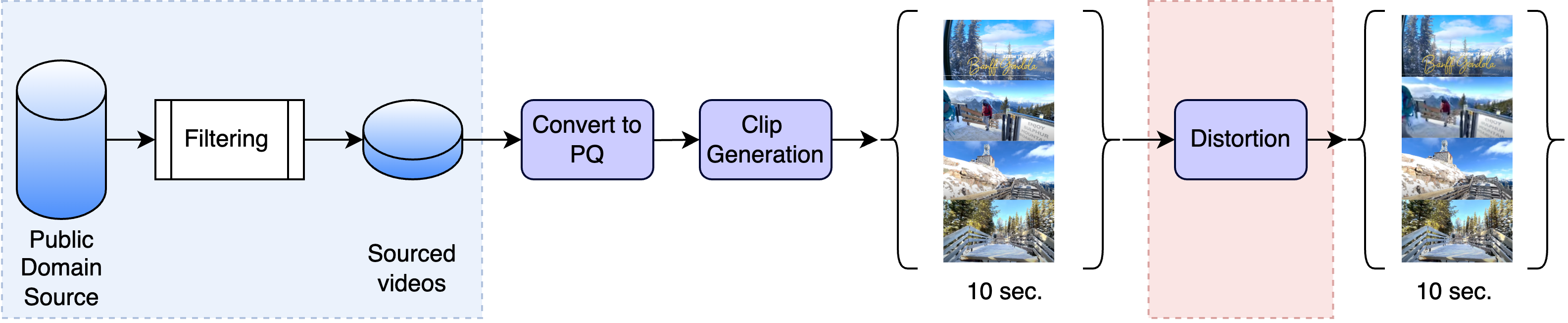}
  \caption{Overview of HDR Fine-tuning database preparation. We obtained the source 4K HDR videos at considerably high bitrates, followed by generating scene-separated 10-second clips and introducing compression and scaling distortions.}
  \label{fig:data1}
\end{figure*}%


\begin{figure*}
  \centering
  \includegraphics[width=0.80\textwidth]{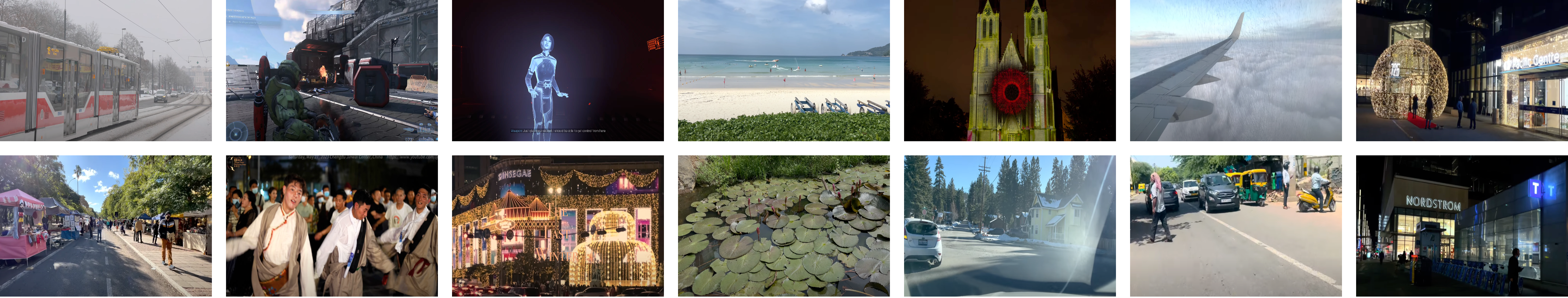}
   \caption{Sample frames from the 411 Source videos depicting content diversity.}
  \label{fig:samplse_video}
\end{figure*}

\begin{figure}
  \centering
  \includegraphics[width=0.8\columnwidth]{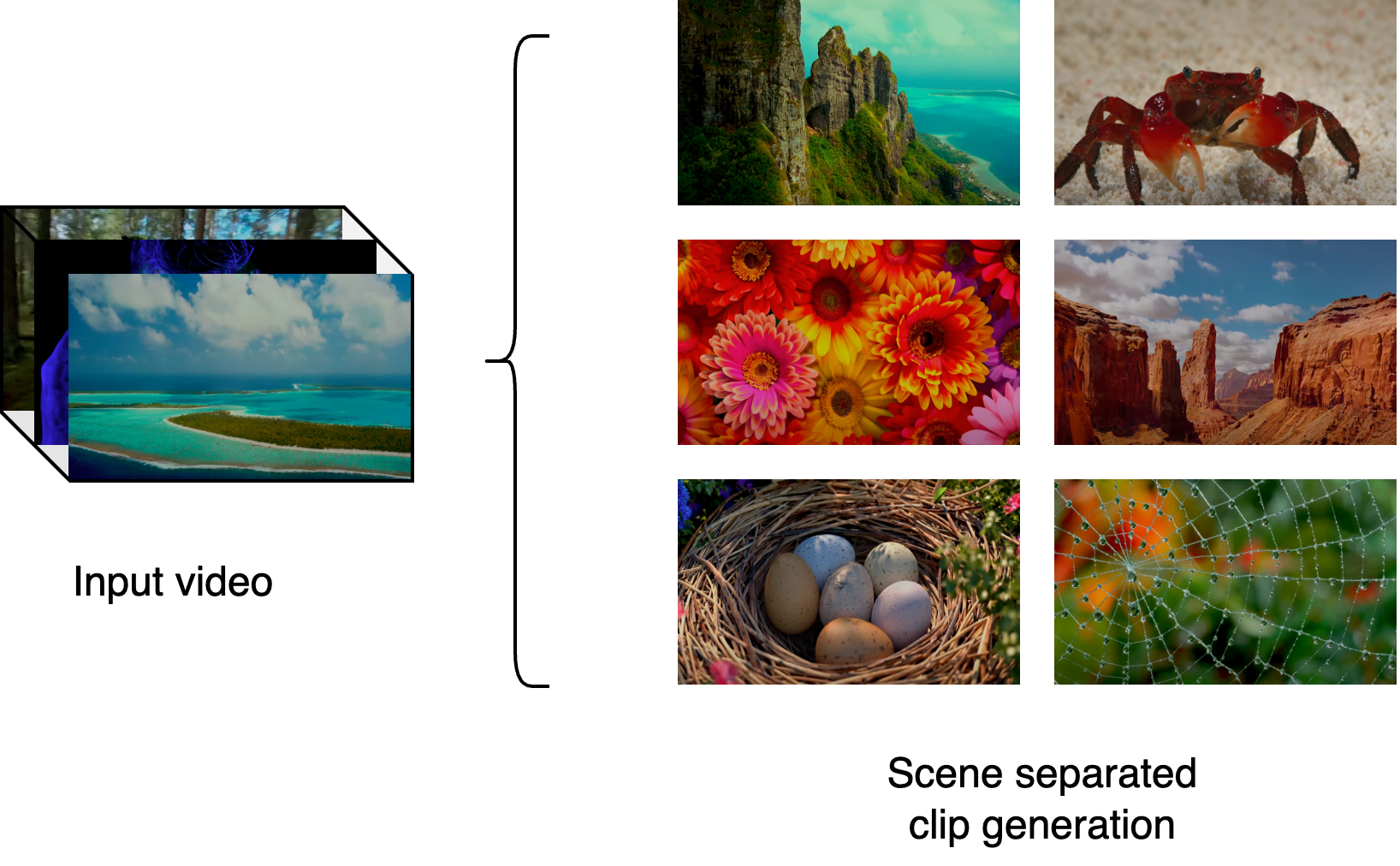}
   \caption{Generation of scene-separated clips from a given HDR video. Each frame on the right is obtained from a different clip.}
  \label{fig:sampleframe}
\end{figure}

\begin{figure}
  \centering
  \includegraphics[width=0.8\columnwidth]{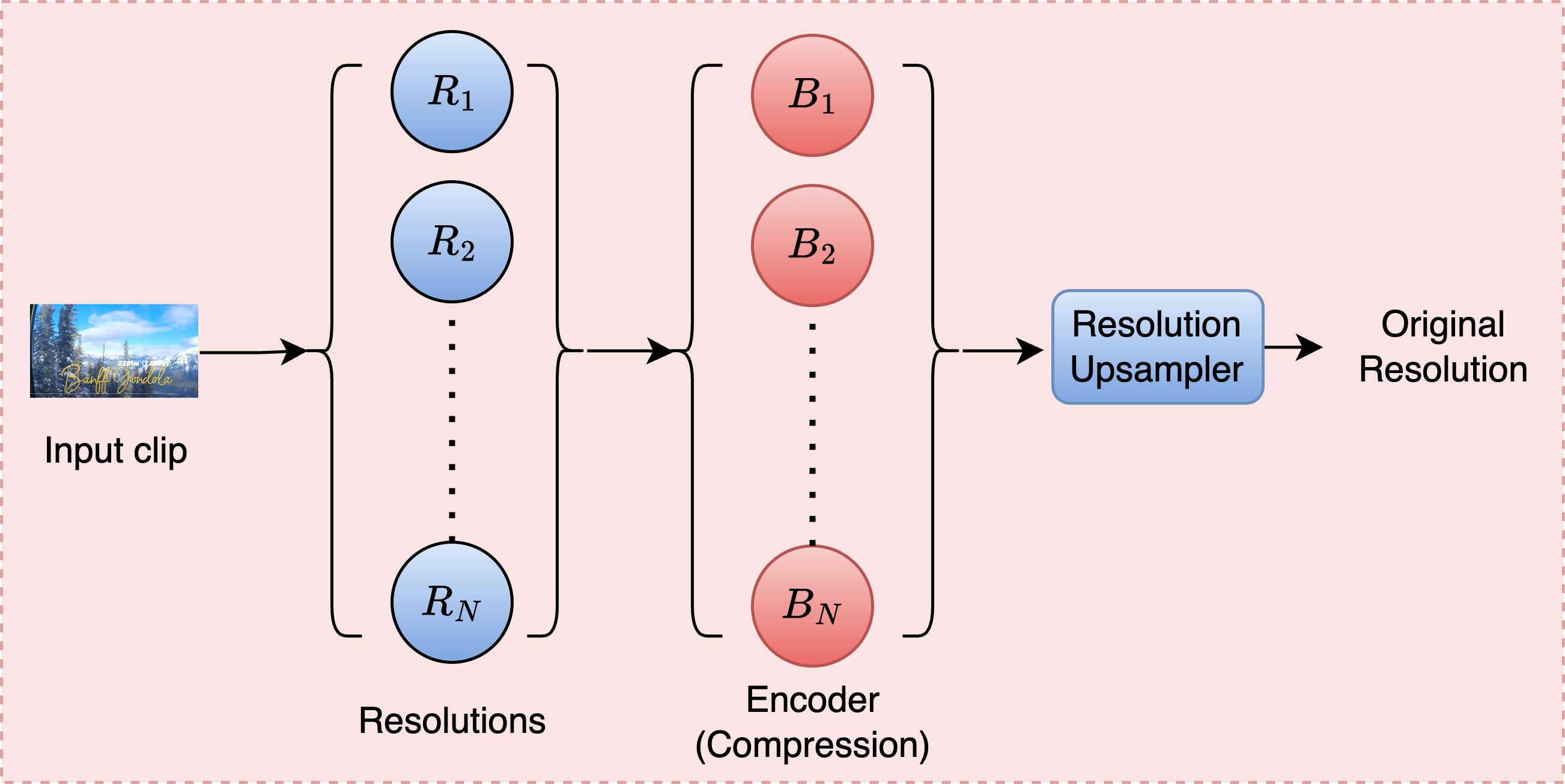}
   \caption{A generic resolution-bitrate ladder creation step followed by a resolution upsampling step to 4K. $R$ and $B$ represent resolution and bitrate, respectively.}
  \label{fig:data2}
\end{figure}

\subsection{HDR Data Collection \& Preparation}
\label{ssec:datacollection}

As access to publicly available HDR databases remains limited, research in the field of HDR quality assessment is still in its early stages. We believe that leveraging unlabeled User-Generated Content (UGC) in HDR format, available in the public domain or under a Creative Commons 4.0 license \cite{CC4}, is a fundamental step for advancing the development and validation of HDR-VQA models. The inherent real-world and diverse characteristics of UGC data provide an added unique advantage toward facilitating the modeling and comprehension of the diverse quality distortions intrinsic to HDR videos.

\begin{table}[]
\centering
\resizebox{0.5\columnwidth}{!}{%
\begin{tabular}{|c|c|}
\hline
Resolution & Bitrate (Mbps) \\ \hline
4K         & 15, 6, 3         \\ \hline
1080p      & 9, 6, 1          \\ \hline
720p       & 4.6, 2.6        \\ \hline
540p       & 2.2            \\ \hline
\end{tabular}%
}
\caption{Bitrate Resolution ladder for Distortions.}
\label{tab:bitladder}
\end{table}
\indent Figure \ref{fig:data1} depicts an overview of our database preparation procedure. Following \cite{hdr-live,hdrtvnet,kay2017kinetics,sr-itm}, we sourced 4K HDR videos. 
These videos were in HDR10 standard, adhering to the Rec.2020 color gamut, with either HLG or PQ OETF, and were 10-bit with a maximum luminance of 1000 nits. Our collection strategy was skewed towards high bitrate ($\approx$ 30 Mbps) videos to preserve pristine quality. We then manually filtered the videos to maintain the content diversity within our dataset. Finally, we collected 411 source videos. Figure \ref{fig:samplse_video} shows example frames from source videos for content diversity. Given the equal split of PQ and HLG transfer functions in the sourced videos, we re-encoded them all to PQ to ensure a consistent data representation format across the dataset and stabilized training. We included videos of varying lengths of more than 4 minutes only. We chose this design based on the assumption that longer videos, which are likely recorded using professional-grade equipment and appropriate HDR settings, would provide a richer and more representative dataset for our study. In contrast, shorter source clips may have been recorded using non-professional-grade equipment and by individuals with limited HDR expertise, potentially offering lower quality and less representative data. \\
\indent We further split the 411 videos into 10-second clips, resulting in a total of 6375 clips. The clipping process was done by segmenting 2-minute continuous scenes, from which a random 10-second clip was extracted. We ensured no overlap occurred between different clips, maintaining the diversity in video scenes of the clips obtained from each source video as shown in Figure \ref{fig:sampleframe}. The final step in our dataset creation involved generating a bitrate-resolution ladder, as per \cite{hdr-live}, to simulate a practical deployment use-case of HDR content on the internet. Figure \ref{fig:data2} shows the bitrate-resolution ladder generation steps. We employed four resolution scales and nine bitrates, as detailed in Table \ref{tab:bitladder}. Note that in addition to distorted videos, we also retained the original pristine clips in the dataset, generating 63750 clips. We believe that this systematic approach in creating the fine-tuning HDR database significantly contributed to the superior performance of the HIDRO-VQA model. 
The hyperlinks to the source videos will be released on GitHub.

\subsection{Self-Supervised SDR Pretraining}
Using self-supervised learning with an auxiliary task offers a feasible route for leveraging the abundant unlabeled data within the video domain, eliminating the need for quality scores. This framework enables the learning of robust and distinctive features from the unlabeled data that are helpful quality-aware representations. \\
\indent As discussed earlier, we use CONTRIQUE as our proposed model's pre-trained SDR quality-aware feature extractor that is fine-tuned with the HDR videos we collected in Section \ref{ssec:datacollection}. CONTRIQUE's architecture consists of two primary parts: an encoder $f(.)$ and a projector $g(.)$. The encoder used is the popular convolutional neural network ResNet-50 \cite{resnet} and focuses on feature extraction, while the projector, a multi-layer perceptron (MLP), reduces the dimensional of the representation from the encoder. CONTRIQUE incorporates multi-scale learning and cropping techniques to capture the inherent multi-scale characteristics of images and their distortions. By analyzing images at two different scales, native resolution, and half-scale resolution, the model gains the ability to capture both local and global image features that are crucial for quality assessment. During training, only two transforms, horizontal flipping and color space conversion, are used. \\
\indent CONTRIQUE assigns a distortion class label to all synthetically distorted images (and their scaled and transformed versions) generated from pristine images with a specific type and degree of distortion. Each image stemming from authentic distortions (UGC type) is treated as a distinct distortion class.  For a given image $x \in \mathbb{R}^{3 \times H\times W}$
\begin{align*}
    h = f(x),  z = g(h) = g(f(x)) \quad h \in \mathbb{R}^D, z \in \mathbb{R}^K
\end{align*}
where h is the D-dimensional output from the encoder, and z is a K-dimensional output from the projector. The dot product is the similarity measure between a pair of representations $\phi(u,v) = u^T v/||u||_2 ||v||_2$. The loss function used in CONTRIQUE is the normalized temperature-scaled cross entropy (NT-Xent), and for image $x_i$ belonging to a synthetically distorted class, it is defined as 
\begin{align}
    \mathcal{L}_i ^{syn} = \frac{1}{|C(i)|} \sum_{j \in C(i)} -\log \frac{\exp(\phi(z_i,z_j)/\tau)}{\sum_{k=1}^N \mathbbm{1}_{k \neq i}\exp(\phi(z_i,z_k)/\tau)}
    \label{eqn:cont_loss_sup}
\end{align}
where $N$ is the total number of images in the batch, $\mathbbm{1}$ is the indicator function, $\tau$ is the temperature parameter, $C(i)$ is a set containing image indices belonging to the same class as $x_i$ (excluding the index $i$) and $|C(i)|$ is its cardinality, and $\tau$ is the temperature parameter.\\
\indent Each UGC image is treated as a unique class. Thus, for a UGC image $x_i$, only its scaled and transformed version $x_j$ belongs to the same class. The loss function for UGC images is given by

\begin{align}
    \mathcal{L}_i ^{UGC} =  -\log \frac{\exp(\phi(z_i,z_j)/\tau)}{\sum_{k=1}^N \mathbbm{1}_{k \neq i}\exp(\phi(z_i,z_k)/\tau)}
    \label{eqn:cont_loss_ugc}
\end{align}

The overall loss function is given by :
\begin{align}
    \mathcal{L} = \frac{1}{N} \sum_{i = 1} ^N \mathbbm{1}_{(x_i \notin UGC)} \mathcal{L}_i ^{syn} + \mathbbm{1}_{(x_i \in UGC)}\mathcal{L}_i ^{UGC}
    \label{eqn:cont_loss_total}
\end{align}
CONTRIQUE is trained on 1.3M images sourced from publicly available databases \cite{kadis,CERTHBLUR,COCO,VOC,AVA}. The framework is trained to handle the diverse mix of unknown distortions present in UGC images as well as synthetic distortions.  \\
\indent 
Our motivation to opt for CONTRIQUE as the pre-trained backbone and fine-tune it using our limited HDR dataset rather than starting from scratch is driven by its demonstrated effectiveness in handling SDR content under various synthetic and real-world distortions. This choice is further validated in our ablation study, as outlined in Section \ref{ssec:sdrpre}. The initial pre-training on SDR data provides a robust foundation, rendering the model well-suited for subsequent fine-tuning on our curated HDR videos. This is also motivated by the fact that quality-aware representations acquired from SDR data can expedite the learning process when working with a comparatively smaller HDR dataset.


\begin{table*}[]
\centering
\resizebox{\textwidth}{!}{%
\begin{tabular}{|cc|ccc|ccc|}
\hline
\multicolumn{2}{|c|}{Viewing Condition} &
  \multicolumn{3}{c|}{Dark Ambient } &
  \multicolumn{3}{c|}{Bright Ambient } \\ \hline
\multicolumn{2}{|c|}{Algorithm} &
  \multicolumn{1}{c|}{SROCC$\uparrow$} &
  \multicolumn{1}{c|}{LCC$\uparrow$} &
  RMSE$\downarrow$ &
  \multicolumn{1}{c|}{SROCC$\uparrow$} &
  \multicolumn{1}{c|}{LCC$\uparrow$} &
  RMSE$\downarrow$ \\ \hline
\multicolumn{1}{|c|}{\multirow{2}{*}{\begin{tabular}[c]{@{}c@{}}Image Quality Metrics using \\ Handcrafted Features\end{tabular}}} &
  HIGRADE \cite{higrade}  &
  \multicolumn{1}{c|}{0.7088 \scriptsize(0.0827)} &
  \multicolumn{1}{c|}{0.6827 \scriptsize(0.0710)} &
  14.2545 \scriptsize(2.0780) &
  \multicolumn{1}{c|}{0.6862 \scriptsize(0.0973)} &
  \multicolumn{1}{c|}{0.6664 \scriptsize(0.0808)} &
  13.7339 \scriptsize(2.0078) \\ \cline{2-8} 
\multicolumn{1}{|c|}{} &
  BRISQUE\cite{brisque} &
  \multicolumn{1}{c|}{0.7251 \scriptsize(0.0955)} &
  \multicolumn{1}{c|}{0.7139 \scriptsize(0.0881)} &
  12.6404 \scriptsize(2.1651) &
  \multicolumn{1}{c|}{0.7133 \scriptsize(0.1004)} &
  \multicolumn{1}{c|}{0.7139 \scriptsize(0.0885)} &
  12.6404 \scriptsize(2.0428) \\ \hline
\multicolumn{1}{|c|}{\multirow{7}{*}{\begin{tabular}[c]{@{}c@{}}Video Quality Metrics using \\ Handcrafted Features/ \\ Supervised Pre-Trained \\ Deep Networks\end{tabular}}} &
  TLVQM \cite{tlvqm} &
  \multicolumn{1}{c|}{0.5781 \scriptsize(0.1014)} &
  \multicolumn{1}{c|}{0.5552 \scriptsize(0.0919)} &
  14.999 \scriptsize(1.9098) &
  \multicolumn{1}{c|}{0.5549 \scriptsize(0.1162)} &
  \multicolumn{1}{c|}{0.5504 \scriptsize(0.1008)} &
  15.2480 \scriptsize(1.8562) \\ \cline{2-8} 
\multicolumn{1}{|c|}{} &
  RAPIQUE \cite{rapique} &
  \multicolumn{1}{c|}{0.4553 \scriptsize(0.2533)} &
  \multicolumn{1}{c|}{0.4864 \scriptsize(0.1171)} &
  15.7134 \scriptsize(1.7415) &
  \multicolumn{1}{c|}{0.4470 \scriptsize(0.2171)} &
  \multicolumn{1}{c|}{0.4910 \scriptsize(0.1393)} &
  15.6088 \scriptsize(1.9382) \\ \cline{2-8} 
\multicolumn{1}{|c|}{} &
  HDR BVQM \cite{hdrbvqm} &
  \multicolumn{1}{c|}{0.6020 \scriptsize(0.0944)} &
  \multicolumn{1}{c|}{0.5844 \scriptsize(0.086)} &
  14.5930 \scriptsize(1.8276) &
  \multicolumn{1}{c|}{0.5411 \scriptsize(0.1102)} &
  \multicolumn{1}{c|}{0.5436 \scriptsize(0.0986)} &
  15.4146 \scriptsize(1.8312) \\ \cline{2-8} 
\multicolumn{1}{|c|}{} &
  VSFA\cite{vsfa} &
  \multicolumn{1}{c|}{0.7127 \scriptsize(0.1079)} &
  \multicolumn{1}{c|}{0.6918 \scriptsize(0.1114)} &
  13.0511 \scriptsize(2.4003) &
  \multicolumn{1}{c|}{0.5549 \scriptsize(0.1162)} &
  \multicolumn{1}{c|}{0.5504 \scriptsize(0.1008)} &
  15.2480 \scriptsize(1.8562) \\ \cline{2-8} 
\multicolumn{1}{|c|}{} &
  V-BLIINDS\cite{vbliinds} &
  \multicolumn{1}{c|}{0.7483 \scriptsize(0.1446)} &
  \multicolumn{1}{c|}{0.7193 \scriptsize(0.1141)} &
  12.7794 \scriptsize(2.3715) &
  \multicolumn{1}{c|}{0.7248 \scriptsize(0.1304)} &
  \multicolumn{1}{c|}{0.7009 \scriptsize(0.1180)} &
  12.896 \scriptsize(2.3606) \\ \cline{2-8} 
\multicolumn{1}{|c|}{} &
  ChipQA \cite{chipqa} &
  \multicolumn{1}{c|}{0.7435 \scriptsize(0.0895)} &
  \multicolumn{1}{c|}{0.7334 \scriptsize(0.0819)} &
  12.1549 \scriptsize(1.9106) &
  \multicolumn{1}{c|}{0.7437 \scriptsize(0.0815)} &
  \multicolumn{1}{c|}{0.7312 \scriptsize(0.0864)} &
  12.3509 \scriptsize(1.843) \\ \cline{2-8} 
\multicolumn{1}{|c|}{} &
  HDR-ChipQA \cite{hdrchipqa} &
  \multicolumn{1}{c|}{0.8250 \scriptsize(0.0589)} &
  \multicolumn{1}{c|}{0.8344 \scriptsize(0.0562)} &
  9.8038 \scriptsize(1.7334) &
  \multicolumn{1}{c|}{0.8316 \scriptsize(0.0580)} &
  \multicolumn{1}{c|}{0.8287 \scriptsize(0.0552)} &
  10.1903 \scriptsize(1.6664) \\ \hline
\multicolumn{1}{|c|}{\multirow{4}{*}{\begin{tabular}[c]{@{}c@{}}Self-Supervised Pre-Trained\\ Image \& Video \\ Quality Algorithms\end{tabular}}} &
  CONTRIQUE  \cite{contrique} &
  \multicolumn{1}{c|}{0.8106 \scriptsize(0.0666)} &
  \multicolumn{1}{c|}{0.7801 \scriptsize(0.0673)} &
  11.5173 \scriptsize(1.8860) &
  \multicolumn{1}{l|}{0.8276 \scriptsize(0.0693)} &
  \multicolumn{1}{l|}{0.7889 \scriptsize(0.0680)} &
  \multicolumn{1}{l|}{11.2970 \scriptsize(2.0186)} \\ \cline{2-8} 

\multicolumn{1}{|c|}{} &
  Re-IQA  \cite{reiqa} &
  \multicolumn{1}{c|}{0.7755 \scriptsize(0.0701)} &
  \multicolumn{1}{c|}{0.7764 \scriptsize(0.1021)} &
  11.1334 \scriptsize(2.6541) &
  \multicolumn{1}{l|}{0.8237 \scriptsize(0.0791)} &
  \multicolumn{1}{l|}{0.7989 \scriptsize(0.1038)} &
  \multicolumn{1}{l|}{10.8499 \scriptsize(2.6619)} \\ \cline{2-8} 

\multicolumn{1}{|c|}{} &
  CONVIQT \cite{conviqt} &
  \multicolumn{1}{c|}{0.8170 \scriptsize(0.0672)} &
  \multicolumn{1}{c|}{0.7875 \scriptsize(0.0705)} &
  11.2514 \scriptsize(2.0548) &
  \multicolumn{1}{l|}{0.8184 \scriptsize(0.0694)} &
  \multicolumn{1}{l|}{0.7857 \scriptsize(0.0700)} &
  \multicolumn{1}{l|}{11.4064\scriptsize(2.0756)} \\ \cline{2-8} 
\multicolumn{1}{|c|}{} &
  HIDRO-VQA (Ours) &
  \multicolumn{1}{c|}{\textbf{0.8793 \scriptsize (0.0672)}} &
  \multicolumn{1}{c|}{\textbf{0.8678 \scriptsize(0.0643)}} &
  \textbf{8.8743\scriptsize(1.7538)} &
  \multicolumn{1}{l|}{\textbf{0.8930 \scriptsize(0.0548)}} &
  \multicolumn{1}{l|}{\textbf{0.8773 \scriptsize(0.0557)}} &
  \multicolumn{1}{l|}{\textbf{8.7110 \scriptsize(1.7911)}} \\ \hline
\end{tabular}%
}
\caption{Median SROCC, LCC, and RMSE on the LIVE-HDR Database on scores collected under the dark and bright ambient conditions of all the compared NR-VQA algorithms. Standard deviations are shown in parentheses. The best-performing algorithm is bold-faced. Results of Algorithms: HIGRADE, BRISQUE, TLVQM, RAPIQUE, HDR BVQM, VSFA, V-BLIINDS, ChipQA, HDR-ChipQA taken from \cite{hdrchipqa}.}
\label{tab:nrvqatable}
\end{table*}
\subsection{HDR Quality-Aware Contrastive Fine Tuning}
\label{ssec:finetune}

HDR videos encompass a wider range of luminance and color, introducing features and distortions that are either absent or less pronounced in SDR videos. This distinction emphasizes the importance of fine-tuning using HDR data to ensure the model is further adapted to the quality-aware features of HDR content and can be effectively used to evaluate HDR video quality. \\
\indent Leveraging the 63,750 video clips acquired through the data processing detailed in Section \ref{ssec:datacollection}, we learned the HIDRO-VQA, model via contrastive fine-tuning. These video clips were generated by applying a distinct resolution-bitrate distortion to reasonably high-bitrate encoded videos. Following the approach utilized in CONTRIQUE for synthetic distortions, we could assign one of the ten distortion class labels to each video clip. These labels cover nine resolution bitrate distortions and the source clip case. \\
\indent However, we propose to perform contrastive fine-tuning on the spatial frames and then employ a simple mean pooling technique to derive video-level features. To achieve this, we randomly select one frame from each video in our database before the start of the training process and fine-tune the SDR pre-trained CONTRIQUE checkpoint using contrastive loss functions. Given the non-uniform distribution of bit allocation across frames in videos during video compression, the resulting frames from the video clips cannot be attributed to one of the ten distortion classes. Consequently, we assume that each frame exhibits a unique type of distortion, similar to the UGC-specific distortion in CONTRIQUE. Thus, our  fine-tuning objective for each frame can be expressed as follows:
\begin{align}
    \mathcal{L}_i ^{HDR-Frame} =  -\log \frac{\exp(\phi(z_i,z_j)/\tau)}{\sum_{k=1}^N \mathbbm{1}_{k \neq i}\exp(\phi(z_i,z_k)/\tau)},
    \label{eqn:cont_ft_loss}
\end{align}
where $z_i$ and $z_j$ represent the outputs of the projector MLP for $x_i$ and $x_j$, which correspond to the input frame and its scaled and transformed counterpart. During the fine-tuning process, we adopt a simplified approach by only applying horizontal flips and the multi-scale feature extraction methodology used in CONTRIQUE.  \\

\section{Experiments \& Results}\label{experiment}
\subsection{HDR Fine-tuning  Configurations}
\label{ssec:training}
The HDR Fine-Tuning was conducted with a batch size set to 768, achieved by selecting frames from the videos obtained in Section \ref{ssec:datacollection}. The selected frames were subsequently cropped to a resolution of 256x256. To extract patch features, we selected a patch size of 64x64, resulting in 4 non-overlapping patches per frame. For each patch, the resultant feature was computed using an adaptive average pooling layer at the end of the ResNet-50 encoder. The temperature parameter ($\tau$) was set to 0.1. The model was fine-tuned for 25 epochs, employing a stochastic gradient descent optimizer with an initial learning rate of 0.1. To ensure stabilized training, the learning rate was subjected to an initial linear warm-up for the first two epochs, followed by a cosine decay schedule without restarts \cite{sgdr}. It should be noted that in our experiment, 25 epochs using only 63750 frames is equivalent to a single epoch of CONTRIQUE training. This shows the data efficiency of our method. The implementations were carried out in Python, making use of the PyTorch framework, and were executed on a workstation equipped with three NVIDIA A100 GPUs.

\subsection{Databases}
We demonstrate the state-of-the-art performance of our model on the LIVE-HDR database. This database consists of 310 videos quality-labeled by human participants under two distinct ambient conditions. The videos were created by applying nine different combinations of compression and downsampling to 31 source videos. The two ambient settings encompassed a dimly lit environment with an incident luminance of less than 10 lux and a well-illuminated setting with an incident luminance of 200 lux. We conducted separate evaluations of HIDRO-VQA on both sets of scores. 

\subsection{Evaluation Protocol}
We utilized a Support Vector Regressor (SVR) with a linear kernel, trained on the features extracted from the fine-tuned ResNet-50 network, to predict the Mean Opinion Scores (MOS) of the videos. Our training protocol involved the following steps: We divided the database into a training set and a test set, maintaining an 80:20 ratio, and ensuring that videos with the same content were exclusively present in one set. This practice aligns with the standard approach for evaluating the performance of Video Quality Assessment (VQA) algorithms and prevents the regressor from capturing content-specific cues. \\
\indent To determine the hyperparameters of the SVR, we conducted a 5-fold cross-validation exclusively on the training set without including videos from the test set. This procedure was iterated 100 times, and the metrics reported reflect the median and standard deviation values. 

\subsection{Performance Metrics}
We evaluated the performance of HIDRO-VQA using three metrics. We calculated Spearman's Rank-Ordered Correlation Coefficient (SROCC) between the predicted scores generated by HIDRO-VQA and the actual ground truth Mean Opinion Scores (MOS). Further, we fit the predicted scores to the MOS using a logistic function
\begin{equation}
l(x) =  \frac{\beta_1-\beta_2}{ 1 + \exp(-\frac{(x - \beta_3)}{ \beta_4}) + \beta_5} 
\end{equation}
and then calculated Pearson's Linear Correlation Coefficient (LCC) and Root Mean Square Error (RMSE) between the fitted scores and the MOS, following the standard practice in the evaluation of VQA algorithms \cite{logistic}.

\subsection{Quantitative Results}
We conducted a performance evaluation of popular NSS-based NR-VQA models, including BRISQUE \cite{brisque}, HIGRADE \cite{higrade}, TLVQM \cite{tlvqm}, V-BLIINDS \cite{vbliinds}, HDR BVQM \cite{hdrbvqm}, VSFA \cite{vsfa}, RAPIQUE \cite{rapique}, ChipQA \cite{chipqa}, and HDR-ChipQA \cite{hdrchipqa}, alongside the popular self-supervised SDR pre-trained models such as CONTRIQUE, Re-IQA, and CONVIQT, using the LIVE-HDR database. In Table \ref{tab:nrvqatable}, we compared the results obtained from our newly introduced model, HIDRO-VQA, with the performance of existing models in our analysis. We adopted the evaluation strategy outlined in \cite{hdrchipqa} for all the NSS-based models. For CONTRIQUE, Re-IQA, and CONVIQT, we converted the videos in the LIVE-HDR database from Y'CbCr to R'G'B', followed by scaling the pixel values to the range [0,1] before feature extraction using publicly accessible checkpoints.\\
\indent The results in Table \ref{tab:nrvqatable} indicate that among the NSS-based models, HIGRADE, BRISQUE, V-BLIINDS, and ChipQA obtained similar performance levels. RAPIQUE, a state-of-the-art NR-VQA model, shows notably poor performance on the LIVE-HDR database, potentially attributed to the resizing of frames performed during feature extraction with its ImageNet pre-trained model. TLVQM, another popular NR-VQA algorithm, exhibits less-than-ideal performance on the HDR content, which might be attributed to the extensive fine-tuning of parameters specific to SDR VQA databases, resulting in difficulties with generalization when applied to HDR databases. An intriguing observation from our analysis reveals that the quality-pretrained deep models, CONTRIQUE, CONVIQT, and Re-IQA, demonstrate remarkable generalization capabilities despite never being trained on HDR databases. Our proposed model, HIDRO-VQA, which uses contrastive fine-tuning of CONTRIQUE on HDR videos, outperformed all the other models we evaluated, enhancing the benchmark on the LIVE-HDR database by 5\%. HIDRO-VQA also achieved a narrower range of SRCC, LCC, and RMSE values compared to most of the compared algorithms, demonstrating its reliability across test sets.

\section{Ablation Studies}\label{ablation}
\subsection{Effect on Number of Fine-Tuning Epochs}
This section presents a comprehensive analysis of the NR-VQA performance achieved by the proposed model, HIDRO-VQA, using varying numbers of fine-tuning epochs.  Our approach involved training HIDRO-VQA over four numbers of distinct epochs: 10, 20, 25, and 30, using the training configurations described in Section \ref{ssec:training}. The results, summarized in Table \ref{tab:epoch}, unveil a significant and consistent trend in performance. We observed continuous and substantial improvement as the number of epochs increased, with significant improvements seen up to 25 epochs. Beyond 25 epochs, the performance plateaued and did not improve on further increasing the fine-tuning epochs. This empirical analysis highlights the pivotal role of determining the optimal number of fine-tuning epochs to achieve the highest level of NR-VQA while optimizing training costs.

\begin{table}[]
\centering
\resizebox{\columnwidth}{!}{%
\begin{tabular}{|cl|ccc|ccc|}
\hline
\multicolumn{2}{|c|}{Viewing Condition} &
  \multicolumn{3}{c|}{Dark Ambient } &
  \multicolumn{3}{c|}{Bright Ambient } \\ \hline
\multicolumn{2}{|c|}{\begin{tabular}[c]{@{}c@{}}HDR Contrastive\\  Finetuning Epochs\end{tabular}} &

  \multicolumn{1}{c|}{SROCC$\uparrow$} &
  \multicolumn{1}{c|}{LCC$\uparrow$} &
  RMSE$\downarrow$ &
  \multicolumn{1}{c|}{SROCC$\uparrow$} &
  \multicolumn{1}{c|}{LCC$\uparrow$} &
  RMSE$\downarrow$ \\ \hline
\multicolumn{2}{|c|}{10} &
  \multicolumn{1}{c|}{0.6383} &
  \multicolumn{1}{c|}{0.6064} &
  14.4268 &
  \multicolumn{1}{c|}{0.6974} &
  \multicolumn{1}{c|}{0.6809} &
  13.4858 \\ \hline
\multicolumn{2}{|c|}{20} &
  \multicolumn{1}{c|}{0.8266} &
  \multicolumn{1}{c|}{0.8123} &
   10.6680 & 
  \multicolumn{1}{c|}{0.8432} &
  \multicolumn{1}{c|}{0.8357} &
   10.0781\\ \hline
\multicolumn{2}{|c|}{25} &
  \multicolumn{1}{c|}{0.8793} &
  \multicolumn{1}{c|}{0.8678} &
  8.8743 &
  \multicolumn{1}{c|}{0.8930} &
  \multicolumn{1}{c|}{0.8773} &
  8.7110 \\ \hline
  \multicolumn{2}{|c|}{30} &
  \multicolumn{1}{c|}{0.8765} &
  \multicolumn{1}{c|}{0.8692} &
  8.8834 &
  \multicolumn{1}{c|}{0.8945} &
  \multicolumn{1}{c|}{0.8798} &
  8.7254 \\ \hline
\end{tabular}%
}
\caption{Number of Fine-tuning Epochs vs. Median SROCC, LCC, and RMSE on the LIVE-HDR Database on scores collected under the dark and bright ambient conditions. }
\label{tab:epoch}
\end{table}

\subsection{Effect of SDR Pre-Training}
\label{ssec:sdrpre}
Furthermore, we explore the impact of pre-training on SDR content on the final performance of our proposed model. To illustrate the efficacy of employing pre-trained SDR models in enhancing the final performance of HDR-VQA, we initiated the training of HIDRO-VQA with randomly initialized model weights instead of using weights from CONTRIQUE  as described in Section \ref{ssec:finetune}. The training followed the configurations detailed in Section \ref{ssec:training}. The results, presented in Table \ref{tab:pretran}, provide evidence of the substantial positive influence of SDR quality-aware pre-training on the final performance. This demonstrates the marked improvement achieved by incorporating SDR content pre-training in the context of HDR-VQA.

\begin{table}[]
\centering
\resizebox{\columnwidth}{!}{%
\begin{tabular}{|cl|ccc|ccc|}
\hline
\multicolumn{2}{|c|}{Viewing Condition} &
  \multicolumn{3}{c|}{Dark Ambient} &
  \multicolumn{3}{c|}{Bright Ambient} \\ \hline
\multicolumn{2}{|c|}{SDR Pretrained?} &
  \multicolumn{1}{c|}{SROCC$\uparrow$} &
  \multicolumn{1}{c|}{LCC$\uparrow$} &
  RMSE$\downarrow$ &
  \multicolumn{1}{c|}{SROCC$\uparrow$} &
  \multicolumn{1}{c|}{LCC$\uparrow$} &
  RMSE$\downarrow$ \\ \hline
\multicolumn{2}{|c|}{No} &
  \multicolumn{1}{c|}{0.3460} &
  \multicolumn{1}{c|}{0.3166} &
  17.1745 & 
  \multicolumn{1}{c|}{0.3325} &
  \multicolumn{1}{c|}{0.2939} &
  17.2928 \\ \hline
\multicolumn{2}{|c|}{Yes} &
  \multicolumn{1}{c|}{0.8793} &
  \multicolumn{1}{c|}{0.8678} &
  8.8743 &
  \multicolumn{1}{c|}{0.8930} &
  \multicolumn{1}{c|}{0.8773} &
  8.7110 \\ \hline
\end{tabular}%
}
\caption{Effect of SDR Pre-Training on Median SROCC, LCC, and RMSE on the LIVE-HDR Database on scores collected under the dark and bright ambient conditions.}
\label{tab:pretran}
\end{table}

\section{Extension to Full Reference VQA}\label{FR-VQA}
Our proposed HIDRO-VQA framework is a flexible approach for obtaining general representations in a Full Reference VQA setting. We refer to this model as HIDRO-FR. We refrained from conducting any additional fine-tuning specifically for FR-VQA, opting to directly employ the learned representations obtained through the fine-tuning process described in Section \ref{ssec:finetune}. To apply these learned representations to FR-VQA, we adopt a straightforward approach. We computed the absolute difference between the features of reference and distorted videos. The representations for the reference and distorted videos were obtained by average pooling the frame feature representations along the temporal dimension, similar to NR-VQA. Like the NR-VQA approach, we employed a Support Vector Regressor (SVR) for the regression task, mapping the obtained representation from the reference and test video to the corresponding Differential Mean Opinion Score (DMOS). \\
\indent Table \ref{tab:frvqa} shows the FR-VQA performance of HIDRO-VQA, along with other state-of-the-art methods. Our evaluation protocol follows the one in \cite{hdrmax}, and we report Spearman's Rank Order Correlation Coefficient (SROCC) and Linear Correlation Coefficient for all the compared methods. In HDRMAX \cite{hdrmax}, the authors introduced a non-linear transformation to transform the luminance values. This enhancement was designed to improve the performance of SDR VQA algorithms for HDR quality assessment. We compare the performances of these algorithms with HIDRO-FR. From Table \ref{tab:frvqa}, it can be observed that HIDRO-FR achieves superior performance compared to other FR-VQA models without using the transformations proposed in \cite{hdrmax}. This underscores the adaptability and broad applicability of our proposed approach. 

\begin{table}[]
\centering
\resizebox{0.7\columnwidth}{!}{%
\begin{tabular}{|c|cc|}
\hline
Dataset          & \multicolumn{2}{c|}{LIVE-HDR (Dark Ambient)}                                                                 \\ \hline
Algorithm        & \multicolumn{1}{c|}{SRCC}                                         & PLCC                                     \\ \hline
HDRMAX \cite{hdrmax}          & \multicolumn{1}{c|}{0.7681 \scriptsize(0.0913)} & 0.7400 \scriptsize(0.0958) \\ \hline
PSNR             & \multicolumn{1}{c|}{0.6242 \scriptsize(0.1504)}     & 0.6357 \scriptsize(0.1331) \\ \hline
PSNR+HDRMAX      & \multicolumn{1}{c|}{0.8263 \scriptsize(0.0684)}     & 0.8206 \scriptsize(0.0615) \\ \hline
SSIM   \cite{ssim}           & \multicolumn{1}{c|}{0.5208\scriptsize(0.1611)}     & 0.4898\scriptsize(0.1595) \\ \hline
SSIM+HDRMAX      & \multicolumn{1}{c|}{0.7771 \scriptsize(0.0866)}     & 0.7529 \scriptsize(0.0964) \\ \hline
MS-SSIM  \cite{mssim}        & \multicolumn{1}{c|}{0.6007 \scriptsize(0.1228)}     & 0.5810 \scriptsize(0.1260) \\ \hline
MS-SSIM+HDRMAX   & \multicolumn{1}{c|}{0.7645 \scriptsize(0.0838)}     & 0.7258 \scriptsize(0.0868) \\ \hline
ST-RRED \cite{strred}         & \multicolumn{1}{c|}{0.6863 \scriptsize(0.0700)}     & 0.6569 \scriptsize(0.0744) \\ \hline
ST-RRED+HDRMAX   & \multicolumn{1}{c|}{0.7896 \scriptsize(0.0607)}     & 0.7595 \scriptsize(0.0603) \\ \hline
SpEED-QA  \cite{speed}        & \multicolumn{1}{c|}{0.6110 \scriptsize(0.1243)}      & 0.6196 \scriptsize(0.1066) \\ \hline
SpEED-QA+HDRMAX  & \multicolumn{1}{c|}{0.7581 \scriptsize(0.0921)}     & 0.7107 \scriptsize(0.0993) \\ \hline
VMAF   \cite{vmaf}          & \multicolumn{1}{c|}{0.6753 \scriptsize(0.0493)}     & 0.6086 \scriptsize(0.0583) \\ \hline
VMAF+HDRMAX & \multicolumn{1}{c|}{0.8528 \scriptsize(0.0543)} & {0.8342 \scriptsize(0.0632)} \\ \hline
HDR-VDP-2 \cite{hdrvdp2}        & \multicolumn{1}{c|}{0.7041 \scriptsize(0.1198)}     & 0.6722 \scriptsize(0.1081) \\ \hline
HDR-VDP-2+HDRMAX & \multicolumn{1}{c|}{0.7431 \scriptsize(0.0770)}     & 0.7208 \scriptsize(0.0764) \\ \hline
HIDRO-FR (ours)    & \multicolumn{1}{c|}{\textbf{0.8699 \scriptsize(0.0388)}}                               &    \textbf{0.8519 \scriptsize(0.0452)} \\ \hline
\end{tabular}%
}
\caption{Median SROCC and LCC were obtained using FR-VQA models. Standard deviations are shown in parentheses. The best-performing algorithm is bold-faced. Results of all algorithms except HIDRO-FR taken from \cite{hdrmax}.}
\label{tab:frvqa}
\end{table}

\section{Conclusion}\label{future}
In this research endeavor, we have introduced a deep learning-based No-Reference Video Quality Assessment (NR-VQA) algorithm tailored to the specific demands of HDR content. We proposed a self-supervised contrastive fine-tuning methodology using unlabeled HDR videos. Our findings underscore the potential of self-supervised pre-trained neural networks initially designed for SDR content to undergo further refinement in a self-supervised context using limited HDR content, culminating in state-of-the-art performance as evidenced by results on the publicly accessible LIVE-HDR VQA database. Although our model achieves state-of-the-art performance by a high margin, there is a scope for improvement. For example, efficient extraction of temporal quality-aware features could further increase VQA performance, and a large-scale database could further benefit the HDR-VQA research community. The source code for HIDRO-VQA will be available on GitHub.

\section{Change Log}
\begin{itemize}
    \item v1: First Upload to arXiv on 18th Nov 2023
    \item v2: Fixed CONTRIQUE NR-VQA result typo in standard deviation values, updated captions for Table 2,5. 
\end{itemize}


\section*{Acknowledgment}
The authors thank the Texas Advanced Computing Center (TACC) at UT Austin for providing compute infrastructure that contributed to the research outcomes in this paper.


{\small
\bibliographystyle{ieee_fullname}
\bibliography{egbib}

\begin{thebibliography}{10}\itemsep=-1pt

\bibitem{CC4}
Creative commons — attribution 4.0 international — cc by 4.0.
\newblock \url{https://creativecommons.org/licenses/by/4.0/}.
\newblock Accessed: 2023-10-23.

\bibitem{hdr-bvqm}
Naima Aamir, Junaid Mir, Imran~Fareed Nizami, Furqan Shaukat, and Muhammad Majid.
\newblock Hdr-bvqm: High dynamic range blind video quality model.
\newblock {\em Multimedia Tools and Applications}, 80:27701--27715, 2021.

\bibitem{hdrbvqm}
Naima Aamir, Junaid Mir, Imran~Fareed Nizami, Furqan Shaukat, and Muhammad Majid.
\newblock {HDR-BVQM:} high dynamic range blind video quality model.
\newblock {\em Multimedia Tools Appl.}, pages 1--15, 2021.

\bibitem{speed}
Christos~G Bampis, Praful Gupta, Rajiv Soundararajan, and Alan~C Bovik.
\newblock {SpEED-QA}: Spatial efficient entropic differencing for image and video quality.
\newblock {\em IEEE Signal Process. Letters}, 24(9):1333--1337, 2017.

\bibitem{norvdpnet}
Francesco Banterle, Alessandro Artusi, Alejandro Moreo, and Fabio Carrara.
\newblock {NOR-VDPNET:} a no-reference high dynamic range quality metric trained on hdr-vdp 2.
\newblock In {\em 2020 IEEE International Conference on Image Processing (ICIP)}, pages 126--130. IEEE, 2020.

\bibitem{hlg}
Tim Borer and Andrew Cotton.
\newblock A display-independent high dynamic range television system.
\newblock {\em SMPTE Motion Imaging Journal}, 125(4):50--56, 2016.

\bibitem{bt.2100}
ITU BT.2100.
\newblock Image parameter values for high dynamic range television for use in production and international programme exchange.
\newblock In {\em Technical Report}. International Telecommunication Union, 2018.

\bibitem{bt.709}
ITU BT.709.
\newblock Parameter values for the {HDTV} standards for production and international programme exchange.
\newblock In {\em Technical Report}. International Telecommunication Union, 2011.

\bibitem{hdrtvnet}
Xiangyu Chen, Zhengwen Zhang, Jimmy~S Ren, Lynhoo Tian, Yu Qiao, and Chao Dong.
\newblock A new journey from {SDRTV to HDRTV}.
\newblock In {\em Proceedings of the IEEE/CVF International Conference on Computer Vision}, pages 4500--4509, 2021.

\bibitem{gamival}
Yu-Chih Chen, Avinab Saha, Chase Davis, Bo Qiu, Xiaoming Wang, Rahul Gowda, Ioannis Katsavounidis, and Alan~C Bovik.
\newblock {GAMIVAL:} video quality prediction on mobile cloud gaming content.
\newblock {\em IEEE Signal Processing Letters}, 30:324--328, 2023.

\bibitem{gru}
Kyunghyun Cho, Bart Van~Merri{\"e}nboer, Caglar Gulcehre, Dzmitry Bahdanau, Fethi Bougares, Holger Schwenk, and Yoshua Bengio.
\newblock Learning phrase representations using {RNN} encoder-decoder for statistical machine translation.
\newblock {\em arXiv preprint arXiv:1406.1078}, 2014.

\bibitem{chipqa}
Joshua~Peter Ebenezer, Zaixi Shang, Yongjun Wu, Hai Wei, Sriram Sethuraman, and Alan~C Bovik.
\newblock {ChipQA}: No-reference video quality prediction via space-time chips.
\newblock {\em IEEE Trans. Image Process.}, 30:8059--8074, 2021.

\bibitem{hdrchipqa}
Joshua~P Ebenezer, Zaixi Shang, Yongjun Wu, Hai Wei, Sriram Sethuraman, and Alan~C Bovik.
\newblock {HDR-ChipQA:} no-reference quality assessment on high dynamic range videos.
\newblock {\em arXiv preprint arXiv:2304.13156}, 2023.

\bibitem{hdrmax}
Joshua~P. Ebenezer, Zaixi Shang, Yongjun Wu, Hai Wei, Sriram Sethuraman, and Alan~C. Bovik.
\newblock Making video quality assessment models robust to bit depth.
\newblock {\em IEEE Signal Processing Letters}, 30:488--492, 2023.

\bibitem{VOC}
Mark Everingham, Luc Van~Gool, Christopher~KI Williams, John Winn, and Andrew Zisserman.
\newblock The pascal visual object classes (voc) challenge.
\newblock {\em International journal of computer vision}, 88:303--338, 2010.

\bibitem{hdrcie}
Mark~D Fairchild and David~R Wyble.
\newblock {HDR-CIELAB} and {HDR-IPT:} simple models for describing the color of high-dynamic-range and wide-color-gamut images.
\newblock In {\em Color Imaging Conference}, volume 2010, pages 322--326, 2010.

\bibitem{live-qualcomm}
Deepti Ghadiyaram, Janice Pan, Alan~C Bovik, Anush~Krishna Moorthy, Prasanjit Panda, and Kai-Chieh Yang.
\newblock In-capture mobile video distortions: A study of subjective behavior and objective algorithms.
\newblock {\em IEEE Transactions on Circuits and Systems for Video Technology}, 28(9):2061--2077, 2017.

\bibitem{rotation_contrastive}
Spyros Gidaris, Praveer Singh, and Nikos Komodakis.
\newblock Unsupervised representation learning by predicting image rotations.
\newblock {\em arXiv preprint arXiv:1803.07728}, 2018.

\bibitem{resnet}
Kaiming He, Xiangyu Zhang, Shaoqing Ren, and Jian Sun.
\newblock Deep residual learning for image recognition.
\newblock 2015.

\bibitem{knovid-1k}
Vlad Hosu, Franz Hahn, Mohsen Jenadeleh, Hanhe Lin, Hui Men, Tam{\'a}s Szir{\'a}nyi, Shujun Li, and Dietmar Saupe.
\newblock The konstanz natural video database (konvid-1k).
\newblock In {\em 2017 Ninth international conference on quality of multimedia experience (QoMEX)}, pages 1--6. IEEE, 2017.

\bibitem{iphone}
iPhone15.
\newblock Apple inc. 2023.
\newblock \url{https://www.apple.com/iphone-15/}.
\newblock Accessed: 2023-10-23.

\bibitem{kay2017kinetics}
Will Kay, Joao Carreira, Karen Simonyan, Brian Zhang, Chloe Hillier, Sudheendra Vijayanarasimhan, Fabio Viola, Tim Green, Trevor Back, Paul Natsev, et~al.
\newblock The kinetics human action video dataset.
\newblock {\em arXiv preprint arXiv:1705.06950}, 2017.

\bibitem{musiq}
Junjie Ke, Qifei Wang, Yilin Wang, Peyman Milanfar, and Feng Yang.
\newblock {MUSIQ:} multi-scale image quality transformer.
\newblock In {\em Proceedings of the IEEE/CVF International Conference on Computer Vision}, pages 5148--5157, 2021.

\bibitem{sr-itm}
Soo~Ye Kim, Jihyong Oh, and Munchurl Kim.
\newblock Deep sr-itm: Joint learning of super-resolution and inverse tone-mapping for {4KUHD HDR} applications.
\newblock In {\em Proceedings of the IEEE/CVF international conference on computer vision}, pages 3116--3125, 2019.

\bibitem{tlvqm}
Jari Korhonen.
\newblock Two-level approach for no-reference consumer video quality assessment.
\newblock {\em IEEE Trans. Image Process.}, 28(12):5923--5938, 2019.

\bibitem{higrade}
Debarati Kundu, Deepti Ghadiyaram, Alan~C Bovik, and Brian~L Evans.
\newblock No-reference quality assessment of tone-mapped {HDR} pictures.
\newblock {\em IEEE Trans. Image Process.}, 26(6):2957--2971, 2017.

\bibitem{eye_luma}
T Kunkel, S Daly, S Miller, and J Froehlich.
\newblock Perceptual design for high dynamic range systems.
\newblock In {\em High Dynamic Range Video}, pages 391--430. Elsevier, 2016.

\bibitem{larsson2017colorization}
Gustav Larsson, Michael Maire, and Gregory Shakhnarovich.
\newblock Colorization as a proxy task for visual understanding.
\newblock In {\em Proceedings of the IEEE conference on computer vision and pattern recognition}, pages 6874--6883, 2017.

\bibitem{vsfa}
Dingquan Li, Tingting Jiang, and Ming Jiang.
\newblock Quality assessment of in-the-wild videos.
\newblock In {\em Proceedings of the 27th ACM International Conference on Multimedia}, MM '19, page 2351–2359, New York, NY, USA, 2019. Association for Computing Machinery.

\bibitem{kadis}
Hanhe Lin, Vlad Hosu, and Dietmar Saupe.
\newblock Kadid-10k: A large-scale artificially distorted {IQA} database.
\newblock In {\em 2019 Eleventh International Conference on Quality of Multimedia Experience (QoMEX)}, pages 1--3. IEEE, 2019.

\bibitem{COCO}
Tsung-Yi Lin, Michael Maire, Serge Belongie, James Hays, Pietro Perona, Deva Ramanan, Piotr Doll{\'a}r, and C~Lawrence Zitnick.
\newblock Microsoft coco: Common objects in context.
\newblock In {\em Computer Vision--ECCV 2014: 13th European Conference, Zurich, Switzerland, September 6-12, 2014, Proceedings, Part V 13}, pages 740--755. Springer, 2014.

\bibitem{sgdr}
Ilya Loshchilov and Frank Hutter.
\newblock {SGDR:} stochastic gradient descent with warm restarts.
\newblock {\em arXiv preprint arXiv:1608.03983}, 2016.

\bibitem{contrique}
Pavan~C Madhusudana, Neil Birkbeck, Yilin Wang, Balu Adsumilli, and Alan~C Bovik.
\newblock Image quality assessment using contrastive learning.
\newblock {\em IEEE Transactions on Image Processing}, 31:4149--4161, 2022.

\bibitem{conviqt}
Pavan~C Madhusudana, Neil Birkbeck, Yilin Wang, Balu Adsumilli, and Alan~C Bovik.
\newblock {CONVIQT:} contrastive video quality estimator.
\newblock {\em IEEE Transactions on Image Processing}, 2023.

\bibitem{ythfr}
Pavan~C. Madhusudana, Xiangxu Yu, Neil Birkbeck, Yilin Wang, Balu Adsumilli, and Alan~C. Bovik.
\newblock Subjective and objective quality assessment of high frame rate videos.
\newblock {\em {IEEE} Access}, 9:108069--108082, 2021.

\bibitem{hdrvdp2}
Rafa{\l} Mantiuk, Kil~Joong Kim, Allan~G Rempel, and Wolfgang Heidrich.
\newblock {HDR-VDP-2}: A calibrated visual metric for visibility and quality predictions in all luminance conditions.
\newblock {\em ACM Trans. Graphics (TOG)}, 30(4):1--14, 2011.

\bibitem{hdrvdp}
Rafal Mantiuk, Karol Myszkowski, and H-P Seidel.
\newblock Visible difference predicator for high dynamic range images.
\newblock In {\em 2004 IEEE International Conference on Systems, Man and Cybernetics (IEEE Cat. No. 04CH37583)}, volume~3, pages 2763--2769. IEEE, 2004.

\bibitem{CERTHBLUR}
Eftichia Mavridaki and Vasileios Mezaris.
\newblock No-reference blur assessment in natural images using fourier transform and spatial pyramids.
\newblock In {\em 2014 IEEE International Conference on Image Processing (ICIP)}, pages 566--570. IEEE, 2014.

\bibitem{brisque}
Anish Mittal, Anush~Krishna Moorthy, and Alan~Conrad Bovik.
\newblock No-reference image quality assessment in the spatial domain.
\newblock {\em IEEE Trans. Image Process.}, 21(12):4695--4708, 2012.

\bibitem{NIQE}
Anish Mittal, Rajiv Soundararajan, and Alan~C Bovik.
\newblock Making a “completely blind” image quality analyzer.
\newblock {\em IEEE Signal processing letters}, 20(3):209--212, 2012.

\bibitem{DIIVINE}
Anush~Krishna Moorthy and Alan~Conrad Bovik.
\newblock Blind image quality assessment: From natural scene statistics to perceptual quality.
\newblock {\em IEEE transactions on Image Processing}, 20(12):3350--3364, 2011.

\bibitem{AVA}
Naila Murray, Luca Marchesotti, and Florent Perronnin.
\newblock Ava: A large-scale database for aesthetic visual analysis.
\newblock In {\em 2012 IEEE conference on computer vision and pattern recognition}, pages 2408--2415. IEEE, 2012.

\bibitem{vmaf}
Netflix.
\newblock {\em VMAF: The Journey Continues}, 2018 (accessed January 13, 2023).

\bibitem{itu-hdrstudy}
Yukihiro Nishida, Amir Nafez, Paul Gardiner, and Andy Quested.
\newblock {ITU-R} study group 6 progress report.
\newblock {\em SMPTE Motion Imaging Journal}, 128(8):70--75, 2019.

\bibitem{CVD2014}
Mikko Nuutinen, Toni Virtanen, Mikko Vaahteranoksa, Tero Vuori, Pirkko Oittinen, and Jukka H{\"a}kkinen.
\newblock {CVD2014—A} database for evaluating no-reference video quality assessment algorithms.
\newblock {\em IEEE Transactions on Image Processing}, 25(7):3073--3086, 2016.

\bibitem{pathak2016context}
Deepak Pathak, Philipp Krahenbuhl, Jeff Donahue, Trevor Darrell, and Alexei~A Efros.
\newblock Context encoders: Feature learning by inpainting.
\newblock In {\em Proceedings of the IEEE conference on computer vision and pattern recognition}, pages 2536--2544, 2016.

\bibitem{imagenet}
Olga Russakovsky, Jia Deng, Hao Su, Jonathan Krause, Sanjeev Satheesh, Sean Ma, Zhiheng Huang, Andrej Karpathy, Aditya Khosla, Michael Bernstein, et~al.
\newblock Imagenet large scale visual recognition challenge.
\newblock {\em International journal of computer vision}, 115:211--252, 2015.

\bibitem{vbliinds}
Michele~A Saad, Alan~C Bovik, and Christophe Charrier.
\newblock Blind prediction of natural video quality.
\newblock {\em IEEE Trans. Image Process.}, 23(3):1352--1365, 2014.

\bibitem{livemcg}
Avinab Saha, Yu-Chih Chen, Chase Davis, Bo Qiu, Xiaoming Wang, Rahul Gowda, Ioannis Katsavounidis, and Alan~C. Bovik.
\newblock Study of subjective and objective quality assessment of mobile cloud gaming videos.
\newblock {\em IEEE Transactions on Image Processing}, 32:3295--3310, 2023.

\bibitem{reiqa}
Avinab Saha, Sandeep Mishra, and Alan~C Bovik.
\newblock {Re-IQA:} unsupervised learning for image quality assessment in the wild.
\newblock In {\em Proceedings of the IEEE/CVF Conference on Computer Vision and Pattern Recognition}, pages 5846--5855, 2023.

\bibitem{vqa-journey}
Avinab Saha, Sai~Karthikey Pentapati, Zaixi Shang, Ramit Pahwa, Bowen Chen, Hakan~Emre Gedik, Sandeep Mishra, and Alan~C Bovik.
\newblock Perceptual video quality assessment: The journey continues!
\newblock {\em Frontiers in Signal Processing}, 3:1193523.

\bibitem{CIE}
J{\'a}nos Schanda.
\newblock {\em {Colorimetry: Understanding the CIE System}}.
\newblock John Wiley \& Sons, 2007.

\bibitem{hdr-live}
Zaixi Shang, Joshua~P Ebenezer, Alan~C Bovik, Yongjun Wu, Hai Wei, and Sriram Sethuraman.
\newblock Subjective assessment of high dynamic range videos under different ambient conditions.
\newblock In {\em 2022 IEEE International Conference on Image Processing (ICIP)}, pages 786--790. IEEE, 2022.

\bibitem{livestream}
Zaixi Shang, Joshua~Peter Ebenezer, Yongjun Wu, Hai Wei, Sriram Sethuraman, and Alan~C. Bovik.
\newblock Study of the subjective and objective quality of high motion live streaming videos.
\newblock {\em IEEE Transactions on Image Processing}, 31:1027--1041, 2022.

\bibitem{logistic}
Hamid~R Sheikh, Muhammad~F Sabir, and Alan~C Bovik.
\newblock A statistical evaluation of recent full reference image quality assessment algorithms.
\newblock {\em IEEE Trans. Image Process.}, 15(11):3440--3451, 2006.

\bibitem{live-vqc}
Zeina Sinno and Alan~Conrad Bovik.
\newblock Large-scale study of perceptual video quality.
\newblock {\em IEEE Transactions on Image Processing}, 28(2):612--627, 2018.

\bibitem{strred}
Rajiv Soundararajan and Alan~C Bovik.
\newblock Video quality assessment by reduced reference spatio-temporal entropic differencing.
\newblock {\em IEEE Transactions on Circuits and Systems for Video Technology}, 23(4):684--694, 2012.

\bibitem{pq}
SMPTE Standard.
\newblock High dynamic range electro-optical transfer function of mastering reference displays.
\newblock {\em SMPTE ST}, 2084(2014):11, 2014.

\bibitem{nits}
Joseph~C Stevens and Stanley~S Stevens.
\newblock Brightness function: Effects of adaptation.
\newblock {\em JOSA}, 53(3):375--385, 1963.

\bibitem{rapique}
Zhengzhong Tu, Xiangxu Yu, Yilin Wang, Neil Birkbeck, Balu Adsumilli, and Alan~C Bovik.
\newblock {RAPIQUE:} rapid and accurate video quality prediction of user generated content.
\newblock {\em IEEE Open Journal of Signal Processing}, 2:425--440, 2021.

\bibitem{yt-ugc}
Yilin Wang, Sasi Inguva, and Balu Adsumilli.
\newblock Youtube ugc dataset for video compression research.
\newblock In {\em 2019 IEEE 21st International Workshop on Multimedia Signal Processing (MMSP)}, pages 1--5. IEEE, 2019.

\bibitem{ssim}
Zhou Wang, A.C. Bovik, H.R. Sheikh, and E.P. Simoncelli.
\newblock Image quality assessment: from error visibility to structural similarity.
\newblock {\em IEEE Transactions on Image Processing}, 13(4):600--612, 2004.

\bibitem{mssim}
Zhou Wang, Eero~P Simoncelli, and Alan~C Bovik.
\newblock Multiscale structural similarity for image quality assessment.
\newblock In {\em The Thrity-Seventh Asilomar Conference on Signals, Systems \& Computers, 2003}, volume~2, pages 1398--1402. Ieee, 2003.

\bibitem{dvl2021}
Fengchuang Xing, Yuan-Gen Wang, Hanpin Wang, Jiefeng He, and Jinchun Yuan.
\newblock {DVL2021:} an ultra high definition video dataset for perceptual quality study.
\newblock {\em Journal of Visual Communication and Image Representation}, 82:103374, 2022.

\bibitem{hosa}
Jingtao Xu, Peng Ye, Qiaohong Li, Haiqing Du, Yong Liu, and David Doermann.
\newblock Blind image quality assessment based on high order statistics aggregation.
\newblock {\em IEEE Transactions on Image Processing}, 25(9):4444--4457, 2016.

\bibitem{CORNIA}
Peng Ye, Jayant Kumar, Le Kang, and David Doermann.
\newblock Unsupervised feature learning framework for no-reference image quality assessment.
\newblock In {\em 2012 IEEE conference on computer vision and pattern recognition}, pages 1098--1105. IEEE, 2012.

\bibitem{lsvq}
Zhenqiang Ying, Maniratnam Mandal, Deepti Ghadiyaram, and Alan Bovik.
\newblock {Patch-VQ:} 'patching up'the video quality problem.
\newblock In {\em Proceedings of the IEEE/CVF conference on computer vision and pattern recognition}, pages 14019--14029, 2021.

\bibitem{QoE}
Xiangxu Yu, Christos~G Bampis, Praful Gupta, and Alan~Conrad Bovik.
\newblock Predicting the quality of images compressed after distortion in two steps.
\newblock {\em IEEE Transactions on Image Processing}, 28(12):5757--5770, 2019.

\bibitem{zhang2016colorful}
Richard Zhang, Phillip Isola, and Alexei~A Efros.
\newblock Colorful image colorization.
\newblock In {\em Computer Vision--ECCV 2016: 14th European Conference, Amsterdam, The Netherlands, October 11-14, 2016, Proceedings, Part III 14}, pages 649--666. Springer, 2016.

\end{thebibliography}
}

\end{document}